\documentclass{article}
\usepackage{amsmath,amsfonts}
\usepackage{algorithmic}
\usepackage[caption=false,font=normalsize,labelfont=sf,textfont=sf]{subfig}
\usepackage{multirow}
\usepackage{pgf}
\usepackage{url}
\usepackage{graphicx}
\usepackage{booktabs}
\usepackage{algorithm, float}
\usepackage{algorithmic}
\usepackage{arxiv}

\def\BibTeX{{\rm B\kern-.05em{\sc i\kern-.025em b}\kern-.08em
    T\kern-.1667em\lower.7ex\hbox{E}\kern-.125emX}}
\usepackage{balance}

\begin{document}
\title{Probabilistic Attention based on Gaussian Processes for Deep Multiple Instance Learning}

\author{Arne Schmidt, Pablo Morales-Álvarez, Rafael Molina
\thanks{This work has received funding from the European Union’s Horizon 2020 research and innovation programme under the Marie Skłodowska Curie grant agreement No 860627 (CLARIFY Project), from the Spanish  Ministry  of  Science  and Innovation under project PID2019-105142RB-C22, and by FEDER/Junta de Andalucía-Consejería de Transformación Económica, Industria, Conocimiento y Universidades under the project P20\_00286. P. Morales-Álvarez acknowledges funding from the University of Granada postdoctoral program ``Contrato Puente''.

A. Schmidt and R. Molina are with the Department of Computer Science and Artificial Intelligence, University of Granada,
Granada, Spain (email: \{arne, rms\}@decsai.ugr.es). P. Morales-Álvarez is with the Department of Statistics and Operations Research, University of Granada, Granada, Spain.

© 2023 IEEE.  Personal use of this material is permitted.  Permission from IEEE must be obtained for all other uses, in any current or future media, including reprinting/republishing this material for advertising or promotional purposes, creating new collective works, for resale or redistribution to servers or lists, or reuse of any copyrighted component of this work in other works.
}
}

\maketitle
\begin{abstract}
Multiple Instance Learning (MIL) is a weakly supervised learning paradigm that is becoming increasingly popular because it requires less labeling effort than fully supervised methods. 
This is especially interesting for areas where the creation of large annotated datasets remains challenging, as in medicine. 
Although recent deep learning MIL approaches have obtained state-of-the-art results, they are fully deterministic and do not provide uncertainty estimations for the predictions. 
In this work, we introduce the Attention Gaussian Process (AGP) model, a novel probabilistic attention mechanism based on Gaussian Processes for deep MIL.
AGP provides accurate bag-level predictions as well as instance-level explainability, and can be trained end-to-end.
Moreover, its probabilistic nature guarantees robustness to overfitting on small datasets and uncertainty estimations for the predictions. 
The latter is especially important in medical applications, where decisions have a direct impact on the patient's health.
The proposed model is validated experimentally as follows. First, its behavior is illustrated in two synthetic MIL experiments based on the well-known MNIST and CIFAR-10 datasets, respectively. 
Then, it is evaluated in three different real-world cancer detection experiments. AGP outperforms state-of-the-art MIL approaches, including deterministic deep learning ones. 
It shows a strong performance even on a small dataset with less than 100 labels and generalizes better than competing methods on an external test set. Moreover, we experimentally show that predictive uncertainty correlates with the risk of wrong predictions, and therefore it is a good indicator of reliability in practice. Our code is publicly available.
\end{abstract}


\section{Introduction}

Machine learning classification algorithms have achieved excellent results in many different applications \cite{lecun_deep_2015, zhao_object_2019, mahmud_applications_2018, li_survey_2021, tpami}. 
However, these algorithms need large datasets that must be labelled by an expert. 
Such labelling process often becomes the bottleneck in real-world applications.
In the last years, Multiple Instance Learning (MIL) has become a very popular weakly supervised learning paradigm to alleviate this burden. 
In MIL, instances are grouped in bags, and only bag labels are needed to train the model \cite{quellec_multiple-instance_2017}.

MIL is especially interesting for the medical field \cite{melendez_novel_2015, quellec_multiple-instance_2017, campanella_clinical-grade_2019, wu_combining_2021}.
As an example, consider the problem of cancer detection in histopathological images, where the goal is to predict whether a given image contains cancerous tissue or not (binary classification problem).
Since these images are extremely large (in the order of gigapixels), they cannot be completely fed to a classifier (typically a deep neural network).
Therefore, the classical approach is to split the image in many smaller patches and train a classifier at patch level. 
Unfortunately, this requires that an expert pathologist labels \emph{every patch} as cancerous or not, which is a daunting, time-consuming, expensive and error-prone task \cite{ciga2021overcoming}.
In the MIL setting, each image is considered as a bag that contains many different instances (its patches). Importantly, since MIL only requires bag labels, the workload for the pathologist is reduced to labelling each image (and not every single patch).

Different underlying classification methods have been proposed for the MIL problem.
Early approaches relied on traditional methods such as support vector machines \cite{andrews_support_2003}, expectation maximization \cite{zhang_em-dd_2002} or undirected graphs \cite{zhou_multi-instance_2009}. 
In recent years, many approaches are based on deep learning models due to their flexibility and their capacity to learn complex functions \cite{yan_deep_2018, wang_revisiting_2018}.
In particular, the current state-of-the-art is given by an attention-based deep learning model originally introduced in \cite{ilse_attention-based_2018}. 
The idea of attention-based MIL is to predict an attention weight for each instance, which determines its influence on the final bag prediction. 

The attention mechanism has several advantages: it can be trained end-to-end with deep learning architectures because it is differentiable, it provides accurate bag-level predictions, and it provides explainability at instance-level (by looking at the instances with higher attention). Due to its success, it has been adapted and extended several times \cite{li_dual-stream_2021, lu_data-efficient_2021, kong_weakly_2019,  Chen_a_semisupervised_2020, Zhang_making_sense_2020}.
However, all these attention-based MIL approaches are based on \emph{deterministic} transformations (usually one or two fully connected layers to calculate the attention weights). This has several drawbacks, such as the lack of uncertainty estimation and the overfitting to small datasets. These limitations can be addressed by introducing a probabilistic model as a backbone for the MIL attention module. Moreover, such a sound probabilistic treatment leads to better predictive performance, see e.g. \cite{gal_2016_dropout_as, alex_kendall_bayesian_2017, RUIZ2019298, morales2020activation}.

In this work, we introduce a novel probabilistic attention mechanism based on Gaussian processes for MIL, which will be referred to as AGP. It leverages a Gaussian process (GP) to obtain the attention weight for each instance. 
GPs are powerful Bayesian models that can describe flexible functions and provide accurate uncertainty estimation due to their probabilistic nature (their most relevant properties will be reviewed in Section \ref{sec:methodology}). 
Moreover, AGP uses variational inference to ensure a probabilistic treatment of the estimated parameters.
We experimentally evaluate AGP on different datasets, including an illustrative MNIST-based MIL problem, CIFAR-10, and three real-world prostate cancer classification tasks. Specifically, we show that: 1) AGP outperforms state-of-the-art and related MIL methods, 2) the estimated uncertainty can be used to identify which model predictions should be disregarded or double-checked, 3) higher attention is assigned to the most relevant instances of each bag (e.g. cancerous patches in images), and 4) AGP generalizes better than competitors to different datasets (and the estimated uncertainty reflects this extrapolation).

In machine learning literature, some related approaches have used GPs in the context of MIL, such as GPMIL \cite{kim_multiple_2014} or VGPMIL \cite{haussmann_variational_2017}.
These methods rely on a sparse GP for the instance classification, followed by an (approximated) maximum function for the final bag prediction. In contrast to our approach, both GPMIL and VGPMIL focus on instance-level predictions which are later combined for bag predictions, they are limited due to the simplicity of the instance aggregation (approximated max-aggregation), and cannot perform end-to-end training with deep learning models. In \cite{wu_combining_2021} we proposed the combination of a deep learning attention model and VGPMIL, but as a two-stage approach: first, the attention mechanism serves to train the feature extractor; in a second step, the VGPMIL model is applied to the extracted features. Although this approach showed promising results, it did not overcome all the aforementioned limitations of VGPMIL.
In particular, the attention mechanism is discarded after the first training phase and not used at all for the final predictions, so the model cannot fully leverage the advantages of combining an attention mechanism and GPs. So three major challenges remain unsolved: (i) end-to-end training, (ii) uncertainty estimation, and (iii) multiclass classification. Our proposed AGP model provides all three features, as described below.

For a different scenario, time series prediction, the combination of GP and attention has recently shown promising results \cite{chen_attentive_2021}.
For this problem, the GP replaced the final regression layer while the attention weights were calculated deterministically, which is a major difference to our work.
Another existing approach combines an attention mechanism with GPs for channel attention \cite{xie_gpca_2021}. Here, the GPs model correlations in activation maps of convolutional neural networks (CNNs) to estimate the channel attention weights. The channel attention helps improve the CNNs performance for visual tasks. However, to the best of our knowledge, our approach is the first one that estimates the attention weights with GPs in the context of deep MIL. 
Moreover, the novel method fully leverages the strengths of GPs for the attention estimation, including the strong function regression capabilities and uncertainty estimation. Indeed, the proposed AGP model does not only provide accurate predictions, outperforming existing state-of-the-art methods, but also provides an estimation of the uncertainty introduced by the attention. At the same time, our model inherits all positive properties of deterministic attention modules, such as explainability on instance level (as later discussed in section \ref{sec:panda}) and end-to-end training with deep learning feature extractors.

The paper is organized as follows. 
Section \ref{sec:methodology} describes the probabilistic model and inference used by AGP, preceded by the notation and background on the attention mechanism and GPs.
Section \ref{sec:mnist_exp} includes an illustrative and visual MIL experiment using MNIST.
In section \ref{sec:prost_exp}, we carry out three experiments on prostate cancer classification.
We not only report a strong performance of the AGP model, but also analyze the probabilistic predictions for these real-world datasets. Finally, section \ref{sec:conclusions} summarizes the main conclusions.

\section{Methodology} \label{sec:methodology}

\noindent
In this section we present the theoretical framework for AGP. First, we describe some required background,  the MIL notation (section \ref{sec:Multiple_Instance_Learning}), the attention mechanism (section \ref{sec:Attention_MIL}), and the basics on (sparse) Gaussian Processes (section \ref{sec:Sparse_Gaussian_Processes}). 
Then, section \ref{sec:Attention_GP} focuses on the description of AGP, including the probabilistic modelling, the variational inference, and how to make predictions. 

\subsection{Multiple Instance Learning (MIL) for Cancer Classification} \label{sec:Multiple_Instance_Learning}
\noindent 
In the classical MIL setting we assume that instances $\chi \in \mathbb{R}^D$ are grouped into bags $\mathcal{X}_b=\{\chi_{b1}, \chi_{b2}, .., \chi_{bN_b}\}$, where the number of instances $N_b$ in each bag $b$ can vary. 
Notice that this notation is not the most standard in MIL, where $X_b$ is usually used for bags and $x_{bi}$ for instances. However, we will use these letters for the input of the GP in Section \ref{sec:Sparse_Gaussian_Processes}. Therefore, to avoid confusion, we have chosen to use $\mathcal{X}_b$ and $\chi_{bi}$ for bags and instances, respectively. 
Each instance has a (binary) label $y_{bi} \in \{0,1\}$ that remains unknown. The bag label $T_b$ is known, and it is given by:
\begin{align}
    T_b &= 0 \ \Leftrightarrow \ \forall i = 1, ... , N_b,\; y_{bi} = 0, \label{eq:mil_neg}\\
    T_b &= 1 \ \Leftrightarrow \ \exists i \in \{1, ... , N_b\}: y_{bi} = 1. \label{eq:mil_pos}
\end{align}

In cancer classification with histopathological images, the MIL setting considers a Whole Slide Image (WSI) as a bag $\mathcal{X}_b$ and its diagnosis as the bag label $T_b$. Since the complete WSI is too big to be processed by a common convolutional neural network, it is sliced into patches that form the instances.
As MIL only requires bag labels for training, the local annotation of the patches by expert pathologists is not necessary. This provides huge benefits in terms of time and cost of the labeling process.
Apart from the binary classification (cancerous vs. non-cancerous), we are also interested in the cancer class of the WSI. This class can be determined based on the features of the cancerous (positive) patches. 
In this setting, equation \eqref{eq:mil_neg} still applies for non-cancerous (negative) WSIs, while for cancerous WSIs we want to specify the class $T_b=c$, with $c \in \{c_1, c_2, .. , c_K\}$ representing each one of the possible $K$ cancer classes based on the cancerous areas. 

\subsection{Attention Mechanism} \label{sec:Attention_MIL}
\noindent 
As mentioned in the introduction, AGP leverages a probabilistic GP-based attention mechanism to aggregate the information of the different instances in a bag. 
This is inspired by the deterministic attention introduced in \cite{ilse_attention-based_2018}. 
Specifically, the model proposed in \cite{ilse_attention-based_2018} consists of three main components: the feature extractor $f_{fe}$, the attention mechanism, and the final classification layer $f_{cl}$. The feature extractor is given by a convolutional neural network followed by some fully connected layers.
It is used to process each instance, resulting in one feature vector $h_{bi} = f_{fe}({\chi_{bi}})$ per instance, with $h_{bi} \in \mathbb{R}^P$. 
Then, the attention mechanism estimates a \emph{deterministic} attention weight per instance $a_{bi}$ based on these features. Specifically, the used mapping is:
\begin{equation}\label{eq:det_att}
     a_{bi}=\frac{\exp\lbrace {w}^\top \tanh(V h_{bi})\rbrace} {\sum_j \exp\lbrace {w}^\top\tanh(V h_{bj})\rbrace},
\end{equation}
where the vector $w \in \mathbb{R}^{L \times 1}$ and the matrix $V \in \mathbb{R}^{L \times P}$ are optimized during training. 
Finally, the classification is done using the average of the extracted features, weighted by the attention values: 
\begin{equation}
    \hat{T}_b = f_{cl}\left( \textstyle\sum_i h_{bi} a_{bi} \right).
\end{equation}

Our goal is now to replace the deterministic attention mechanism given in equation \eqref{eq:det_att} by a GP model. The GP is a probabilistic method that is able to give a better estimation of the attention weights. Moreover, it allows for capturing the uncertainty introduced by the attention mechanism. 

\subsection{(Sparse) Gaussian Processes}\label{sec:Sparse_Gaussian_Processes}

\noindent 
Gaussian Processes are stochastic processes where the output distribution is assumed to be multivariate Gaussian \cite{rasmussen_gaussian_2006}. 
They can be used to estimate an objective function $f$ in a probabilistic way: the GP defines a prior distribution over functions (whose properties depend on the type of kernel used), and the posterior is computed given such prior and the observed data \cite{morales2017remote}. 
The major drawback of GPs is that the computation of the posterior is not scalable, because it involves inverting a matrix of size $N \times N$, with $N$ the number of datapoints \cite{svendsen2020deep}. 
This is clearly relevant for our MIL scenario, where there exist typically plenty of instances.

To overcome this limitation, different types of Sparse Gaussian Processes (SGPs) have been introduced in the last years \cite{snelson_sparse_2006, hensman_scalable_2015, liu2020gaussian, morales2019scalable}. 
Here we will follow the approach in \cite{hensman_scalable_2015}, since it allows for training in batches. 
The idea behind SGP is to define the GP posterior distribution on a set of $M$ inducing point locations $Z = \{z_i\}_{i=1}^M$, instead of doing it on the $N$ real instances $X = \{x_i\}_{i=1}^N$. The amount of inducing points is taken $M\ll N$, and their location must be representative for the training distribution (in fact, they can be optimized during training, as we will do in AGP). 

The formulation of SGP is as follows. Let $U$ be the output of the GP at the inducing point locations $Z$, and $F$ the output at the datapoints $X$. The SGP model is given by
\begin{align}
    p(U|Z) &= \mathcal{N}(U|0, K_{ZZ}), \label{eq:SGP_1}\\
p(F|U, Z, X) &= \mathcal{N}(F|K_{XZ}K^{-1}_{ZZ}U, \hat{K}), \label{eq:SGP_2}
\end{align}
where $K_{AB}:=k(A,B)$ is the matrix obtained by applying the GP kernel function on $A$ and $B$. Moreover, we have
\begin{equation} \label{eq_k_hat}
    \hat{K} = K_{XX} - K_{XZ} K_{ZZ}^{-1} K_{ZX}. 
\end{equation}
To perform inference, a posterior Gaussian distribution $q(U)=\mathcal{N}(U|\mu_u,\Sigma_u)$ is used on the inducing points. Therefore, the parameters to be estimated during training are $\mu_u$, $\Sigma_u$, the kernel parameters, and the inducing points locations. 

Given a test point, the prediction of the SGP model is a random variable (and not a single deterministic value). The mean of the random variable provides the value to regress, while the standard deviation provides the uncertainty.
Finally, in this section we have written $X$ for the input to the GP, as is standard in the GP literature. 
However, we want to stress that in our case the input to the GP will be given by the features extracted in some previous step, and not the raw input itself.

\subsection{Probabilistic attention based on Gaussian Process (AGP)} \label{sec:Attention_GP}

\begin{figure*}[!t] 
    \centering
    \includegraphics[width=0.8\textwidth]{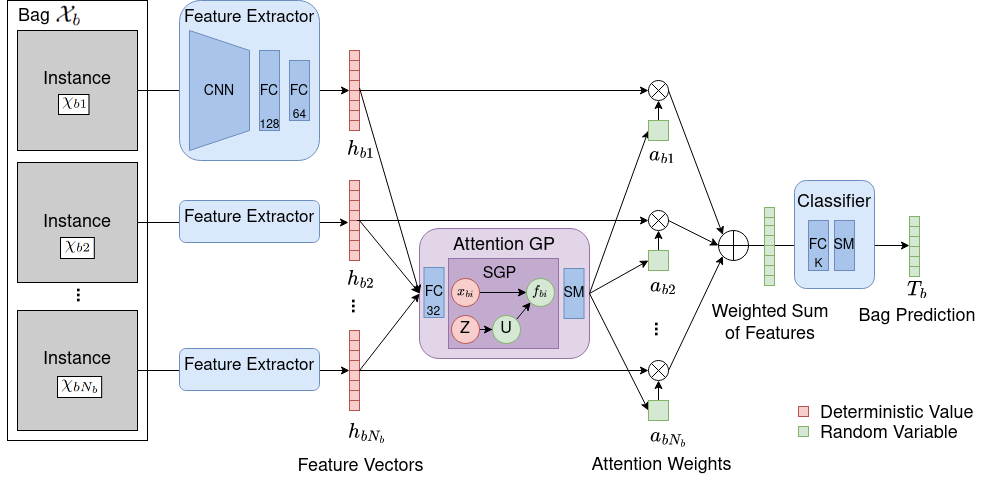}
    \caption{The AGP model architecture. The feature extractor consists of a Convolutional Neural Network (CNN) and two Fully Connected Layers (FC). The attention module incorporates another FC layer, the Sparse Gaussian Process (SGP), and a softmax (SM) function. The final classification is performed by another fully connected layer (where the dimensionality depends on the number of classes) and a softmax activation. While the feature vectors are deterministic values, the use of a (sparse) GP makes the attention weights and the final prediction random variables. 
    } 
    \label{fig:model}
\end{figure*}

\noindent 
\textbf{Probabilistic modelling}.
The AGP model is depicted in Figure \ref{fig:model}. It is a combination of a deterministic convolutional network that serves as a feature extractor $f_{fe}$, an SGP to estimate the attention, and a deterministic fully connected layer for the final classification $f_{cl}$. Next, we describe the different components using Figure \ref{fig:model} as reference.

Remember that $\mathcal{X}_b=\{\chi_{b1}, ..,\chi_{bN_b} \} $ describe the instances in one bag $b$. First, the feature extractor $f_{fe}$ is applied. It consists of a CNN and two fully connected layers with ReLu activation and 128 and 64 units, respectively. The choice of the CNN backbone depends on the task. For cancer classification, we will use EfficientNetB5 as the CNN backbone \cite{tan_efficientnet_2019}. The output of the feature extractor are high-level feature vectors $H_b=\{h_{b1}, ..,h_{bN_b}\}$ where each $h_{bi}, i = 1,..,N_b$ has 64 dimensions:
\begin{equation}
    H_b = f_{fe}({\mathcal{X}_b}).
\end{equation}

We focus now on the attention module. Here, we first apply to each instance the same fully connected layer with sigmoid activation and 32 units. 
This further reduces the dimensionality, resulting in the SGP input feature vectors $X_b=\{x_{b1}, ..,x_{bN_b} \} $ of 32 dimensions each. This alleviates the optimization of the inducing point locations, which are defined in the input space.
 Moreover, the sigmoid function guarantees values between 0 and 1, which facilitates the initialization of the inducing point locations. In summary, each vector $h_{bi}$ with 64 components is transformed into a vector $x_{bi}$ with 32 components.
With these feature vectors, the SGP model described in Section \ref{sec:Sparse_Gaussian_Processes} regresses the values $F_b=\{f_{b1}, ..,f_{bN_b}\}$, which are normalized through a softmax (SM) layer to calculate the attention weights $A_b=\{a_{b1}, ..,a_{bN_b}\}$:
\begin{equation}\label{eq:att_sm}
    a_{bi} = \frac{\exp\lbrace f_{bi}\rbrace}{\sum_j \exp\lbrace f_{bj}\rbrace}.
\end{equation}
Importantly, note that the attention weights are random variables, since they are computed from the SGP output. During inference, we will use Monte Carlo sampling to approximate their distribution.  

Finally, the classifier $f_{cl}$ is used to obtain the bag label $T_b$. The bag label $T_b$ follows a categorical distribution with $K$ classes, $T_b\sim \mathrm{Cat}(p_1,\dots,p_K)$, $\sum_k p_k=1$. These probabilities are computed by applying the classifier $f_{cl}$ over the average of the feature vectors $H_b$ weighted by the attention weights $A_b$: 
\begin{equation}\label{eq:cl}
    (p_1,\dots,p_K) = f_{cl}\left(\textstyle\sum_i h_{bi} a_{bi}\right).
\end{equation}
As shown in Figure \ref{fig:model}, the classifier $f_{cl}$ consists of one fully connected layer with one unit per class and a softmax activation function. Again, since the attention weights $A_b$ are random variables, the probabilities $p_1,\dots,p_K$ are random variables whose distribution will be estimated through Monte Carlo sampling. This probabilistic nature will allow for computing uncertainty estimation in the predictions. 

Once we have described how AGP processes one bag {$\mathcal{X}_b$}, the joint full probabilistic model is
\begin{equation}\label{eq:full_model}
    p(T,F,U) = p(T|F)p(F|U)p(U).
\end{equation}
Here, we have written $T=\{T_1,\dots, T_B\}$ for the collection of all the bag labels, and analogously for $F$. The inducing points $U$ and their locations $Z$ are global for all the bags because the feature space is the same for all instances of all bags and the SGP should be able to generalize to unseen bags.
Notice that, to lighten the notation, we are not writing explicitly the dependency on all the variables.
For instance, $p(U)=p(U|Z)$ depends on the inducing point locations $Z$; $p(F|U)=p(F|U,Z, X)$ also depends on $Z$ and the SGP input $X$; and $p(T|F)=p(T|F,X)$ depends on $X$ and depends on $F$ only through $A$ (recall eqs.~\eqref{eq:att_sm}-\eqref{eq:cl}).
Also, we are not writing explicitly the dependency on other parameters such as all the neural network weights (which are collectively denoted as $W$) and the SGP kernel parameters (which are denoted as $\theta$). 

\textbf{Variational inference}.
To perform inference in AGP, we need to obtain the posterior distribution $p(F,U|T)$ and the learnable parameters $W$, $\theta$, $Z$. 
Since eq.~\eqref{eq:full_model} is not analytically tractable, we resort to variational inference \cite{zhang2018advances}.
Namely, variational inference considers a parametric posterior distribution and finds the parameters that minimize the distance to the true posterior in the Kullback-Leibler divergence sense (by maximizing the log evidence lower bound, ELBO). 
In our case, we consider the distribution $q(F,U) = p(F|U)q(U)$,
where $p(F|U)$ equals the (prior) conditional distribution in eq.~\eqref{eq:SGP_2} and $q(U)=\mathcal{N}(U|\mu_u,\Sigma_u)$ is a (multivariate) Gaussian with mean vector $\mu_u$ and covariance matrix $\Sigma_u$, both to be estimated during training (variational parameters).

With this choice, the ELBO to be maximized is
\begin{equation}\label{eq:ELBO}
     \scalebox{0.9}{$\log p(T) \geq \mathrm{ELBO}=\mathbb{E}_{q(F)} \log p(T|F) - \mathrm{KL}(q(U)||p(U))$},
\end{equation}
where $q(F) = \int p(F|U)q(U) dU$ is a Gaussian distribution since both $p(F|U)$ and $q(U)$ are Gaussian, recall eqs.~\eqref{eq:SGP_1}-\eqref{eq:SGP_2}.
{Specifically, we have $q(F)=\mathcal{N}(F|K_{XZ}K^{-1}_{ZZ}\mu_u,K_{XX} - K_{XZ} K_{ZZ}^{-1}(K_{ZZ}-\Sigma_u)K_{ZZ}^{-1}K_{ZX})$.} 
The KL divergence $\mathrm{KL}(q(U) || p(U))$ can be calculated in closed-form, since both distributions are Gaussian too. In practice, this term acts as a regularizer for the SGP model, since it encourages the posterior on the inducing points to stay close to the prior. To calculate the other term (log-likelihood), we have 
\begin{equation}\label{eq:loglik}
     \mathbb{E}_{q(F)} \log p(T|F) = \textstyle\sum_{b} \mathbb{E}_{q(F_b)} \log p(T_b|F_b), 
\end{equation}
since we naturally assume that bag labels are independent. Although the terms $\mathbb{E}_{q(F_b)} \log p(T_b|F_b)$ cannot be obtained in closed-form, they can be approximated by Monte Carlo integration:
\begin{equation}\label{eq:loglik_term}
    \mathbb{E}_{q(F_b)} \log p(T_b|F_b) \approx  \frac{1}{S}\textstyle\sum_s \log p^{(s)}_{T_b}.
\end{equation}
Here, the subindex $T_b$ indicates that we take the class probability that corresponds to the (observed) bag label $T_b$ (recall from eq.~\eqref{eq:cl} that there exists one $p_k$ for each class). The $S$ samples $\{p^{(s)}_{T_b}\}_{s}$ are obtained by sampling $F_b^{(s)}$ from the Gaussian $q(F_b)$ with the reparametrization trick \cite{kingma_variational_2015} and propagating through the rest of the network (that is, applying eqs.~\eqref{eq:att_sm}-\eqref{eq:cl}). Notice that maximizing the log-likelihood term is equivalent to minimizing the standard cross-entropy between the estimated class probabilities and the ground truth vector (in a one-hot encoding), as shown in \cite{gal_2016_dropout_as}.

In summary, AGP training consists in maximizing the ELBO in eq.~\eqref{eq:ELBO} with respect to the variational parameters ($\mu_u$ and $\Sigma_u$), the neural network parameters $W$, the SGP kernel parameters $\theta$ and the inducing point locations $Z$. To do so, we use stochastic optimization with the Adam algorithm and mini-batches \cite{kingma2014adam}. In the experiments, each mini-batch is given by all the instances of one bag.
Notice that the proposed model and inference allow for end-to-end training through the ELBO maximization.
Algorithm \ref{alg} summarizes the training process.

\begin{algorithm}[tb]
\caption{AGP training procedure} \label{alg}
 \begin{algorithmic}
 \renewcommand{\algorithmicrequire}{\textbf{Input:}}
 \renewcommand{\algorithmicensure}{\textbf{Output:}}
 \REQUIRE Instances $\{\chi_{bi}\}_{i=1,..,N_b}$ (e.g. image patches) for each bag $b=1,\dots,B$; bag labels $\{T_b\}$; number of epochs $E$.
 \ENSURE  Optimal model parameters $Z, \mu_u, \Sigma_u, \theta, W$. \\
  \FOR {$e= 1$ to $E$ (all epochs)}
  \FOR {$b= 1$ to $B$ (all bags)}
  \STATE Predict features $H_b \gets f_{fe}(\mathcal{X}_{b})$.
  \STATE Apply fully connected layer in the attention module, i.e. $X_b \gets f_{FC}(H_{b})$.
  \STATE Calculate SGP output $p(F_b|U,Z,X_b)$ (eq. \ref{eq:SGP_2}).
  \STATE Draw $S$ Monte-Carlo samples $\tilde{F}_{b}^s \sim p(F_{b}|U,Z, X_{b})$.
  \STATE Calculate log likelihood (LL) term following eq.~\eqref{eq:loglik_term}.

  \STATE Calculate KL term in eq.~\eqref{eq:ELBO} in closed-form.
  \STATE Calculate loss as $\mathcal{L}=-\mathrm{LL}+\mathrm{KL}$.
  \STATE Update $Z, \mu_u, \Sigma_u, \theta, W$ with $\nabla \mathcal{L}$ using Adam.
  \ENDFOR
  \ENDFOR
 \RETURN Optimal model parameters $Z, \mu_u, \Sigma_u, \theta, W$.
 \end{algorithmic}
 \end{algorithm}

\noindent
\textbf{Predictions}.
After training is completed, we are interested in predicting the class label $T_b^*$ for a previously unseen bag ${\mathcal{X}_b^*}$. 
The prediction of the AGP model is given by a $K$-class categorical distribution with class scores $p_1,\dots , p_K$ which are random variables. The mean of such random variables, $\overline p_k$, represents the predicted probabilities per class (and the predicted class is the one with highest probability, i.e. $T^*_b=\arg\max_k \overline p_k$). 
Additionally, the standard deviation of each class probability provides its degree of uncertainty. The total uncertainty for the bag prediction is defined as the mean of the standard deviations for each class. 
Notice that this approximation follows popular existing literature \cite{hensman_scalable_2015, gal_2016_dropout_as}.

To calculate the predictive random variables $p_k$, we have 
\begin{equation}
     p(T_b^*) 
     = \textstyle\int p(T_b^*|F_b) q(F_b) d F_b.
\end{equation}
Similarly to eq.~\eqref{eq:loglik}, this cannot be obtained in closed-form, since it requires integrating out the neural network $f_{cl}$. Following the same idea as there, we approximate the predictive distribution by Monte Carlo sampling. Namely, we take $S$ samples from $q(F_b)$ and propagate them through the rest of the network (eqs.~\eqref{eq:att_sm}-\eqref{eq:cl}) to obtain $S$ samples $p_k^{(s)}$ for each class $k=1,\dots,K$. The mean and standard deviation for each $p_k$ are computed empirically based on these samples. 
Analogously, we have $S$ samples $a_{bi}^{(s)}$ for the predictive distribution over the attention weights for each instance inside the bag.
We will see that these values provide explainability on which instances are the most relevant to obtain the bag prediction. 
Using just $S=20$ samples works well in practice. 

\textbf{Implementation}.

Algorithm \ref{alg} provides an overview of the implementation. It summarizes the main steps to train the model. 
In practice, the model is implemented with the deep learning libraries tensorflow (version 2.3.0) and its extension tensorflow-probability (version 0.11.1). 
The class \textit{tensorflow\textunderscore probability.layers.VariationalGaussianProcess()} is specially useful for the implementation of the SGP. It allows for an efficient, parallel execution of the algorithm on the GPU, as well as end-to-end training. Another benefit is the easy integration in other existing deep learning projects.
The complete code for AGP is publicly available. \footnote{\url{https://github.com/arneschmidt/attention_gp}}

Another key component for the implementation is the reparametrization trick to sample from $q(F_b)$, recall eq.~\eqref{eq:loglik_term}. The (Gaussian) output distribution of the SGP is 'reparametrized' into one deterministic part and one probabilistic part to perform backpropagation through the probabilistic layer. Namely, the random vector $F_b \sim \mathcal{N}(\mu, \Sigma)$
can be split into $(\mu + L\epsilon) \sim \mathcal{N}(\mu, \Sigma)$, where $L$ is the Cholesky factor of $\Sigma$ and $\epsilon \sim \mathcal{N}(0, I)$.
For MC sampling, a random sample $\hat{\epsilon} \sim \mathcal{N}(0, I)$ is drawn to obtain a sample from $F_b$, i.e. $\hat{F}_b=\mu + L \hat{\epsilon}$. This allows the backpropagation of the gradient through $\mu$ and $L$, while the random variable $\epsilon$ is independent from the model parameters.
The reparametrization trick is already implemented in the above mentioned tensorflow-probability library. For further details we refer the interested reader to the original work \cite{kingma_variational_2015}.

\section{Synthetic Experiments} \label{sec:synth_exp}

In this section we evaluate our method on two synthetic MIL problems.
First, Section \ref{sec:mnist_exp} shows a visual experiment based on MNIST that helps better understand the proposed method.
Then, Section \ref{sec:cifar_exp} provides a more sophisticated multi-class problem where we compare our method against a wide range of state-of-the-art baselines.

\subsection{An illustrative example: MNIST bags} \label{sec:mnist_exp}
\noindent
The goal of this section is to illustrate the behavior of AGP in a simple and intuitive example. Specifically, we analyze two aspects of AGP: 1) its predictive performance. We will see that bag-level predictions are correct and high attention weights are given to positive instances. 2) The information provided by the estimated uncertainty (i.e. the standard deviation of the predictions). We will see that high uncertainty is assigned to bags that are difficult to classify.

The well-known MNIST dataset \cite{deng_mnist_2012} contains 60000 training and 10000 test images.
To define a MIL problem, we randomly group these images into bags of 9 instances each.
We define the digit ``0'' as the positive class, and the rest of digits as negative class.
Thus, a bag is positive if at least one of the nine digits in that bag is a ``0''. Otherwise, the bag is negative.
We choose ``0'' because it can be mistaken with ``6'' or ``9''. Similar procedures to define a MIL problem on MNIST have been used in previous work \cite{ilse_attention-based_2018}. The resulting MIL dataset has 6667 bags for training (2558 negative, 4109 positive), which are made up of 60.000 instances (54077 negative, 5923 positive). For testing it has 1112 bags (404 negative, 708 positive), which are made up of 10.000 instances (9020 negative, 980 positive). For all splits, around 40\% of the bags have one positive instance, 17\% two, 4\% three and 1\% four and the rest are negative bags.

As the given problem is less complex than the cancer classification task, we simplify the feature extractor. Namely, it contains one convolutional layer (4 filters, 3x3 convolutions) and one fully connected layer (64 units). The rest of the model remains as described in Figure \ref{fig:model}. We train it end-to-end with cross-entropy and the Adam optimizer with a learning rate of 0.0001, for 5 epochs.

\begin{figure}
     \centering
     \subfloat[\label{fig:f_a}]{
         \centering
         \fbox{\includegraphics[width=0.4\linewidth]{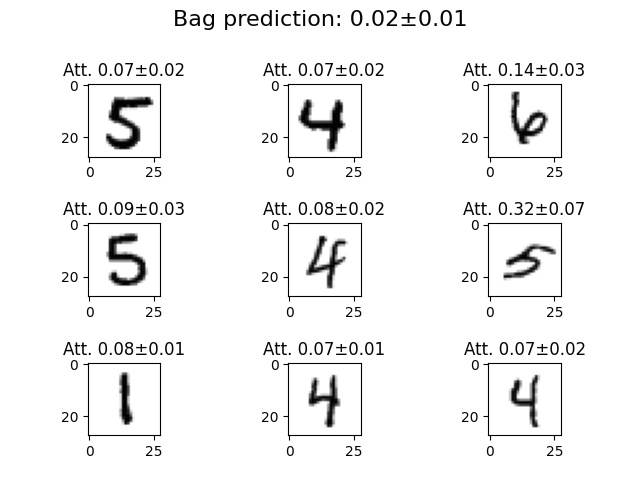}}
     }
     \subfloat[\label{fig:f_b}]{
         \centering
         \fbox{\includegraphics[width=0.4\linewidth]{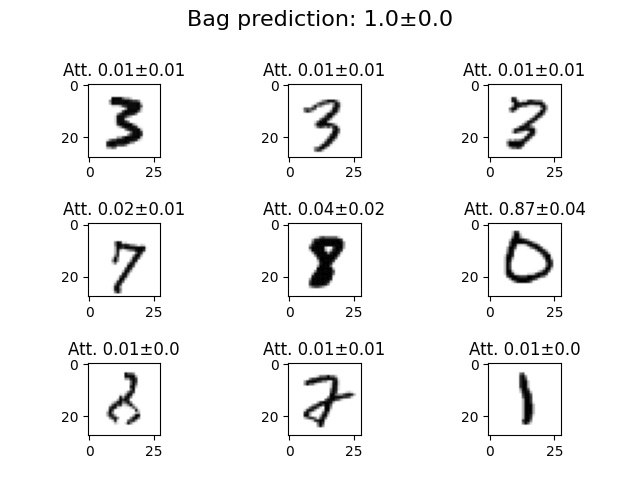}}
     }
    \\
     \subfloat[\label{fig:f_c}]{
         \centering
         \fbox{\includegraphics[width=0.4\linewidth]{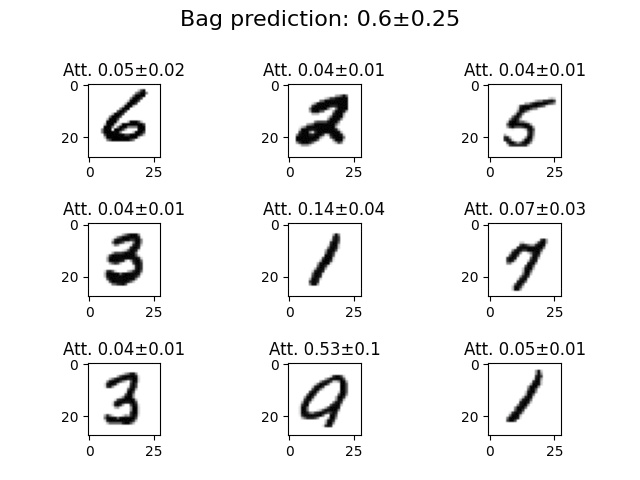}}
     }
     \subfloat[\label{fig:f_d}]{
         \centering
         \fbox{\includegraphics[width=0.4\linewidth]{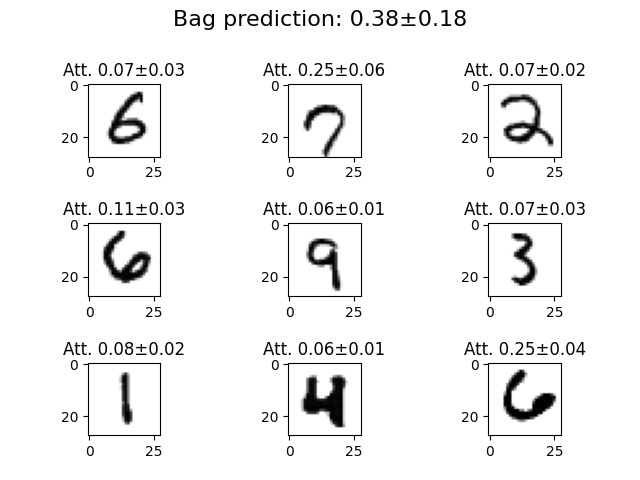}}
     }
        \caption{AGP classification of MNIST bags. In this dataset, the number ``0'' represents a positive instance, and all other digits are considered negative instances. We show the AGP bag prediction (probability to be positive) and corresponding attention weights for each instance. As these estimations are random variables in the AGP model, we report the mean and standard deviation. The top two figures show confident predictions for a negative (a) and a positive (b) bag. The bottom figures, (c) and (d), show two unconfident predictions with high standard deviations. Both bags are negative, but the model misclassifies (c) due to an ambiguous digit.}
        \label{fig:mnist_examples}
\end{figure}

Regarding the predictive performance, AGP achieves $98.02\%$ test accuracy at bag level. This is slightly better than when using deterministic attention (i.e. A-Det, which obtains $97.82\%$). Figures \ref{fig:f_a} and \ref{fig:f_b} show the predictions obtained by AGP for a negative and a positive bag, respectively. 
Notice that AGP learns to discriminate between positive and negative instances, assigning a high attention weight to the digit ``0'' in the positive bag. 
Also, notice that the standard deviation for both the attention weights and the bag prediction is low (i.e. the algorithm is confident on the decision).

Next, we analyze the role of the uncertainty by visualizing the prediction for ambiguous bags. In Figures \ref{fig:f_c} and \ref{fig:f_d} we see two examples of predictions with high standard deviations. These high standard deviations originate in ambiguous instances that lead to an uncertain final bag prediction. In Figure \ref{fig:f_c}, there is a '9' which is visually similar to a '0' because one line is (almost) missing. The model assigns a high attention but also a high standard deviation to this instance. The final bag prediction is false positive, but the high standard deviation of the bag prediction indicates a high uncertainty. In Figure \ref{fig:f_d}, two digits are ambiguous (those corresponding to the items $(1,2)$ and $(3,3)$ of the $3\times 3$ matrix of digits). They are assigned a higher attention but again a slightly higher standard deviation than the other instances. The final negative bag prediction is correct, but the high standard deviation reflects a high uncertainty.
Finally, notice that this qualitative observation on the uncertainty can also be confirmed statistically: while correctly classified bags have an average standard deviation of $0.006$, the average standard deviation of incorrectly classified bags is more than ten times higher ($0.065$). Therefore, a high standard deviation indicates a high risk of a wrong prediction.

\subsection{Evaluation on CIFAR-10} \label{sec:cifar_exp}
{
In this experiment with the CIFAR-10 dataset \cite{cifar10} we want to compare different deterministic and probabilistic approaches for a more difficult multi-class MIL problem.

The CIFAR-10 dataset consists of 32x32 images containing 10 different classes (airplanes, cars, birds, cats, deer, dogs, frogs, horses, ships, and trucks). 
The dataset contains 10.000 test images and we split the remaining images into 50.000 for training and 10.000 for validation. 
Originally, all labels of the images are known, but we create a multi-class MIL problem for our use-case. 
As in the MNIST experiment, we use bags of nine instances each. 
In this case, we select two positive classes while all other classes are negative, as explained next in more detail. 
Each bag has either the label 'airplane', 'car' or 'negative'. 
Negative bags contain only negative instances (i.e. birds, cats, deer, dogs, frogs, horses, ships, and/or trucks). 
Bags labelled as 'airplane' contain at least one image of airplane, while the remaining instances are negative. Similarly, each bag with the label 'car' contains at least one image of cars (and the rest of instances are negative).
We choose to have an equal distribution of each class on the bag-level for training (1481 bags per class, 4443 bags in total), validation (370 bags per class, 1110 bags in total) and test (370 bags per class, 1110 bags in total). As the instances are drawn randomly, the exact amount per class vary in this setup. On average, 5.67\% of the instances are from the 'airplane' class, 5.67\% from the 'car' class, and the rest are negative instances. 

We use the model architecture described in section \ref{sec:methodology} and depicted in Figure \ref{fig:model}.
For the feature extraction, we choose a CNN backbone that consists of 3x3 convolutions with relu activation and max pooling with a stride of 2x2. The exact layers are: two convolutional layers with 32 filters, max pooling, two convolutional layers with 64 filters, max pooling, two convolutional layers with 128 filters and max pooling.
The fully connected layers have 128 and 64 units, respectively. The whole architecture is trained end-to-end. We use the Adam optimizer with a learning rate of 0.0001, cross-entropy, and 15 training epochs.

We compare our method against three state-of-the-art deterministic baselines that only differ in the MIL aggregation mechanism. 
In all the cases, we use the same feature extractor architecture, hyperparameters and iterations. The compared methods are:
}
\begin{itemize}
    \item \textbf{Mean Aggregation (Mean-Agg)}. Instead of using an attention module, we aggregate the extracted features from each instance by taking their mean. This mean vector is then used for the final classification.
    \item \textbf{Attention Deterministic (A-Det)}. The attention module as proposed in \cite{ilse_attention-based_2018} is used. Their attention weights and the final prediction are deterministic values.
    \item \textbf{Attention Deterministic Gated (A-Det-Gated)}. The advanced attention module proposed in \cite{ilse_attention-based_2018} as an extension to A-Det is used. The gating mechanism was introduced to allow the algorithm to efficiently learn more complex relationships between instances.
    \item \textbf{Attention Gaussian Process (AGP)}. The probabilistic model proposed in this work, as described in section \ref{sec:methodology}.
\end{itemize}

\begin{figure}
     \centering
     \includegraphics[width=0.3\linewidth]{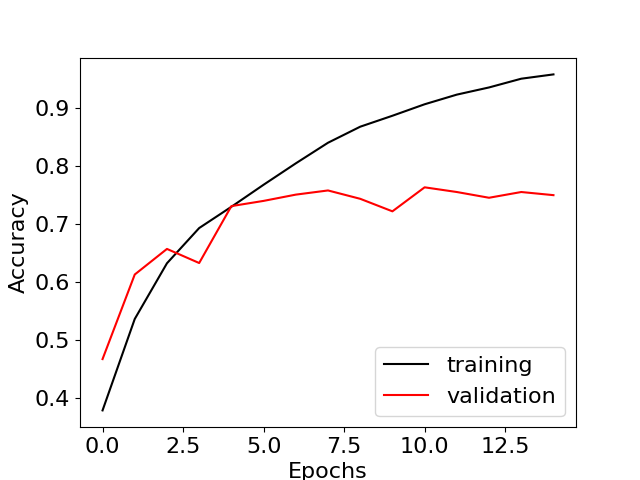}
     \caption{Training and validation accuracy for the AGP model in the CIFAR-10 experiment. Although the amount of labeled bags is low, the model is robust to overfitting as the validation accuracy remains stable.} \label{fig:cifar_train}
\end{figure}

As shown in Table \ref{tab:cifar10}, AGP outperforms all the baselines, including the state-of-the-art attention mechanism A-Det-Gated.  
This suggests that our probabilistic attention is able to accurately assign attention weights to different instances.
Indeed, this can be explained by the good regression capabilities of GPs, as known from previous studies \cite{snelson_sparse_2006, liu2020gaussian, morales2019scalable}.
To illustrate the learning process, we plot a training/validation curve of the AGP model in Figure \ref{fig:cifar_train}. As seen in the plot, the model is robust to overfitting as the validation accuracy remains stable once it converges.

Finally, as a first approach towards future work, we also investigated other options to implement a probabilistic attention mechanism, based on Bayesian Neural Networks (BNNs) instead of the SGP. 
Leaving the rest of the architecture as shown in Figure \ref{fig:model}, we first exchange the AGP attention mechanism by two fully connected Bayesian layers with weights following a Gaussian distribution \cite{Vladimirova2019UnderstandingPI} (32 and 1 units for the layers, respectively). 
In the same experiment setup, this model achieved 0.642 accuracy and 0.644 F1-score, quite far from AGP.
Similarly, we tested a model based on MC dropout \cite{gal_2016_dropout_as} with an attention mechanism composed by one fully connected layer with 32 units, a Bayesian dropout layer, and a fully connected layer with 1 unit. The dropout probability was set to 0.5. This model obtained better results, achieving 0.74 accuracy and 0.739 F1-score. However, this is still lower than AGP (0.749 accuracy, 0.750 F1-score), which will be the focus in terms of probabilistic methods in the rest of this paper.

\begin{table}
\begin{center}
\caption{Results for Cifar-10 experiments with bags of 9 images and three classes. The experiment was repeated in 10 independent runs, we report the mean and standard error.} \label{tab:cifar10}
\label{tab:comparison_cifar}
\begin{tabular}{rccccc}
\toprule
 Method &Type&Acc. mean&Acc. S.E.&F1 mean&F1 S.E.\\
\midrule
Mean-Agg & Det. & 0.732 & 0.008 & 0.730 & 0.009\\
A-Det   &  Det. &  0.735 &  0.003 & 0.735 & 0.003\\
A-Det-Gated   &  Det. & 0.730 &  0.005 & 0.730 & 0.005\\
AGP   & Prob. &    \textbf{0.749} & 0.008 & \textbf{0.750} & 0.008     \\ 
\bottomrule
\end{tabular}
\end{center}
\end{table}

\section{Experiments on prostate cancer classification} \label{sec:prost_exp}
\noindent 
In this section, we evaluate AGP on the real-world problem of cancer classification. This is a very timely problem, since the development of computer aided diagnosis tools is attracting plenty of attention due to the large workload that pathologists are experiencing in the last years \cite{campanella_clinical-grade_2019, lopez2021learning}. For easier reproducibility, we use publicly available datasets. We will focus on prostate cancer, although our model is agnostic to the cancer type and can be applied to other cancer classification tasks.

In the rest of this section, we present the datasets used (SICAPv2 and PANDA), the implementation details, and the baselines used for comparison. Then, Section \ref{sec:sicap} focuses on SICAPv2 data, Section \ref{sec:panda} focuses on PANDA data, and Section \ref{sec:extrapolation} evaluates the ability to extrapolate from one dataset to the other.
The goal of these experiments is not only to show the strong performance of AGP on real-world data, but also to highlight the usefulness of the probabilistic output to estimate the predictive reliability.

\textbf{Datasets.}
We use two publicly available datasets: SICAPv2 and PANDA.
The extracted biopsies (WSIs) are classified by pathologists based on the appearance and quantity of cancerous tissue. There are two different scales: the Gleason Score and the ISUP grade. For further background on both scales, we refer the interested reader to \cite{epstein_2014_2016}. In our experiments, we use the Gleason Score for SICAPv2 and the ISUP grade for PANDA. The reason for this is twofold. First, to compare our results with previous literature (for which we need to use the same grading scale as them). Second, to show that the proposed method is robust to the grading scale used (obtaining good results in both scenarios).

SICAPv2\footnote{Available at: https://data.mendeley.com/datasets/9xxm58dvs3/1}  consists of 155 biopsies (WSIs).
The class distribution of the assigned Gleason Score (GS) is the following: Non-Cancerous: 36, GS6: 14, GS7: 45, GS8: 18, GS9: 35, GS10: 7. The dataset is already split into four cross-validation folds, which contain between 86 and 97 WSIs for training and a separate set for testing. The publishers of the data distributed the biopsies so that the class proportions are reflected in each of the train and test splits. For more details, see \cite{Silva-rodriguez2020GoingDetection}.
The PANDA dataset\footnote{Available at: https://www.kaggle.com/c/prostate-cancer-grade-assessment} consists of 10616 WSIs and was presented at the MICCAI 2020 conference as a challenge. The total number of WSIs for each ISUP grade is the following: Non-Cancerous: 2892, G1: 2666, G2: 1343, G3: 1242, G4: 1249, G5: 1224.
As the test set of PANDA is not publicly available, we use the train/validation/test split proposed in \cite{silva-rodriguez_self-learning_2021}, which has 8469, 353 and 1794 WSIs, respectively. Again, each split follows the overall class proportions.

For both datasets, a 10x magnification is used and the WSIs are split into 512x512 patches with a 50\% overlap. The patch-level annotations of the datasets are discarded for our experiments, since our model only requires bag labels for training.

\textbf{Implementation Details.}
The AGP architecture used for these experiments is depicted in Figure \ref{fig:model}, and explained in Section \ref{sec:Attention_GP}. Here we provide the rest of the details for full reproducibility. We use 64 inducing points with 32 dimensions each, whose locations are initialized with random values between 0.3 and 0.7 (because this is typically the range of values initially obtained by the previous sigmoid layer). For Monte-Carlo integration, we draw 20 samples for training and for testing.
We use a class balanced loss with cross-entropy and the Adam optimization algorithm. We set the learning rate to 0.001 for the first 10 epochs, and use learning rate decay afterwards with the factor $e^{-0.1}$ per epoch. The total number training epochs is set to 100.
Finally, as it is computationally unfeasible to train the feature extractor on all patches of a (huge) WSI at once, we first extract the high-level features of each patch. We train the CNN and the first fully connected layer with the method proposed in \cite{schmidt_coupling}, using only WSI labels. The obtained 128 dimensional feature vectors per patch are then used to train the last fully connected layer of the feature extractor and the rest of the model. 

\textbf{Baselines.}
In all the experiments, we compare mean aggregation as a baseline and three state-of-the-art MIL approaches trained with the same feature vectors, hyperparameters and iterations. The only variation is the attention mechanism. Additionally, in each experiment we compare with other related approaches that have used the same data. For details about the approaches (Mean-Agg, A-Det, A-Det-Gated) please recall the bullet points in Section \ref{sec:cifar_exp}.

\textbf{Evaluation metric}. As common in prostate cancer classification tasks \cite{silva-rodriguez_going_2020, silva-rodriguez_self-learning_2021, arvaniti_automated_2018}, we report the performance in terms of quadratic Cohen's kappa, which measures the agreement between the labels provided by pathologists and the model's predictions. A kappa value of 0 means no agreement (random predictions) and a kappa value of 1 means complete agreement. In all cases, we show the mean and the standard error of the results over several independent runs.

\subsection{SICAPv2 Results}\label{sec:sicap}
\noindent 
\begin{table}
\centering
\caption{Ablation studies with the SICAPv2 dataset. We study the effect of three important components: the feature vector dimension, the number of inducing points for the SGP, and the activation function used inside the attention module. Bold letters indicate the configuration used in the final setup.}\label{tab:abl_study}
{
\begin{tabular}{lccc}
\toprule
 Method      & $\kappa$ mean & $\kappa$ S.E.\\
 \midrule
AGP-feat-dim-32  &    0.832  & 0.004\\
\textbf{AGP-feat-dim-64}  &    \textbf{0.847} & \textbf{0.001}\\
AGP-feat-dim-128  &   0.835 & 0.001\\
AGP-feat-dim-256  &   0.844 & 0.001\\
\midrule
AGP-ind-points-16  &    0.832  & 0.004\\
AGP-ind-points-32  &    0.840  & 0.002\\
\textbf{AGP-ind-points-64}  &    \textbf{0.847} & \textbf{0.001}\\
AGP-ind-points-128  &   0.689 & 0.007\\
\midrule
AGP-relu &   0.675 & 0.008\\
\textbf{AGP-sigmoid} &    \textbf{0.847}  &   \textbf{0.001}\\ 
AGP-tanh &    0.830  &   0.007\\ 
\bottomrule
\end{tabular}
}
\end{table}

\begin{table}
\centering
\caption{Results for SICAPv2 dataset. We report the mean and standard error of Cohen's quadratic kappa ($\kappa$) for 4 independent runs, with a four-fold cross-validation in each run. The last two methods do not report the standard error.\label{tab:comparison_sicap}}
\begin{tabular}{rccc}
\toprule
 Method      & Learning          & $\kappa$ mean & $\kappa$ S.E.\\
 \midrule
Mean-Agg  & MIL &    0.800  & 0.041\\
A-Det  & MIL &    0.770 & 0.008\\
A-Det-Gated  & MIL &   0.814 & 0.007\\
AGP  & MIL &    \textbf{0.847}  &   0.001\\ 
Arvaniti et al. \cite{arvaniti_automated_2018} \cite{silva-rodriguez_going_2020} & Supervised        &    0.769        & N.A.\\
Silva-Rodríguez et al. \cite{silva-rodriguez_going_2020} & Supervised        &    0.818        & N.A.\\
\bottomrule
\end{tabular}
\end{table}

\begin{figure*}
    \centering
     \subfloat[Mean-Agg]{
         \centering
         \includegraphics[trim={1cm 0.0cm 2.5cm 2cm}, width=0.22\linewidth]{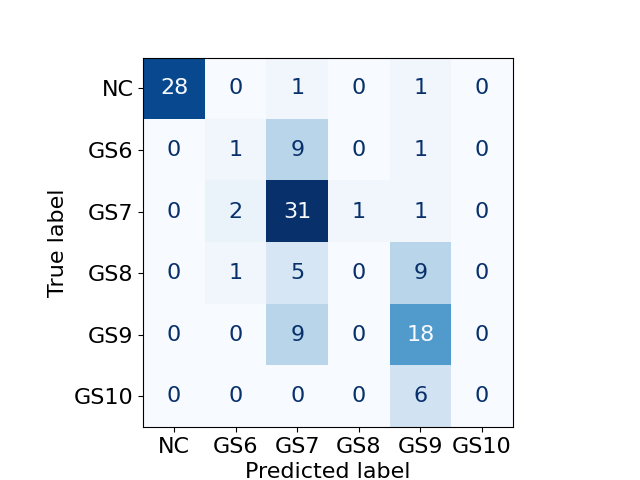}
    }      
    \subfloat[A-Det]{
         \centering
         \includegraphics[trim={1cm 0.0cm 2.5cm 2cm}, width=0.22\linewidth]{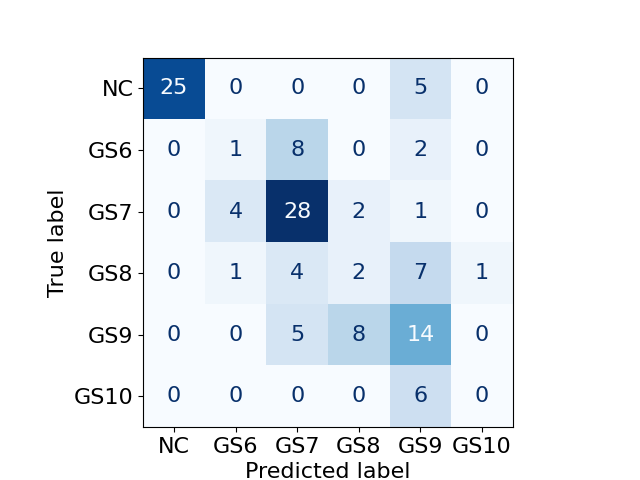}
    }     
     \subfloat[A-Det-Gated]{
         \centering
         \includegraphics[trim={1cm 0.0cm 2.5cm 2cm}, width=0.22\linewidth]{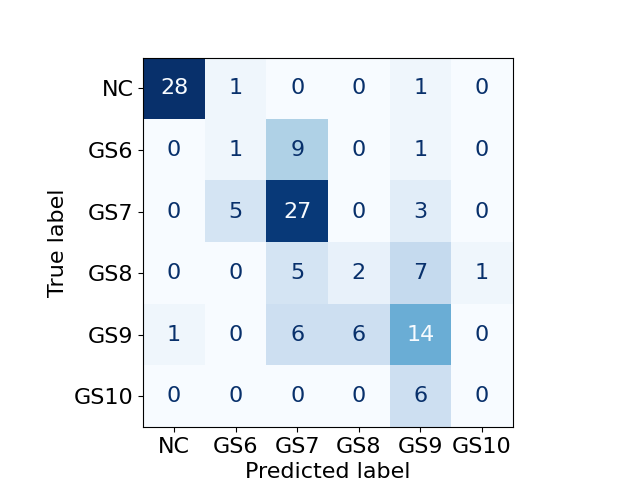}
    }
    
    \bigskip
    
     \subfloat[AGP]{
         \centering
         \includegraphics[trim={1cm 0.0cm 2.5cm 2cm}, width=0.22\linewidth]{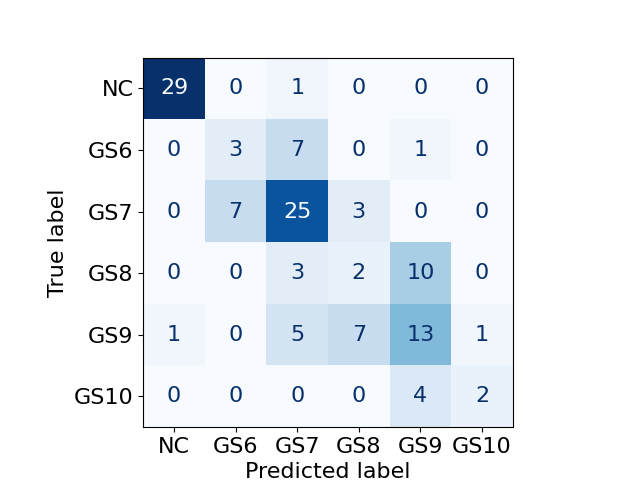}
    }  
    \subfloat[Arvaniti et al.]{
         \centering
         \includegraphics[trim={1cm 0.0cm 2.5cm 2cm}, width=0.22\linewidth]{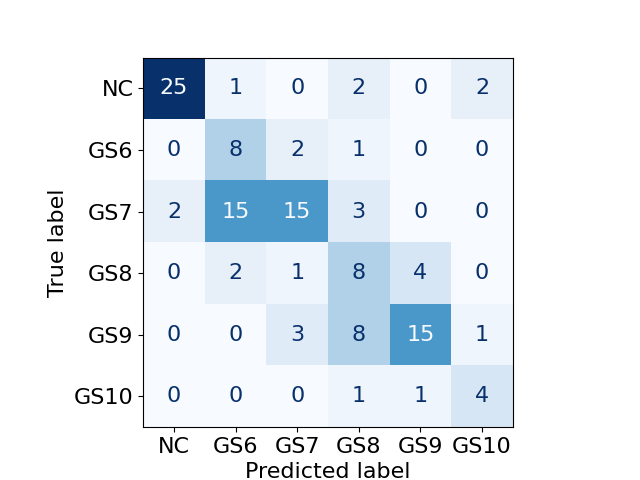}
    }     
     \subfloat[Silva-R. et al.]{
         \centering
         \includegraphics[trim={1cm 0.0cm 2.5cm 2cm}, width=0.22\linewidth]{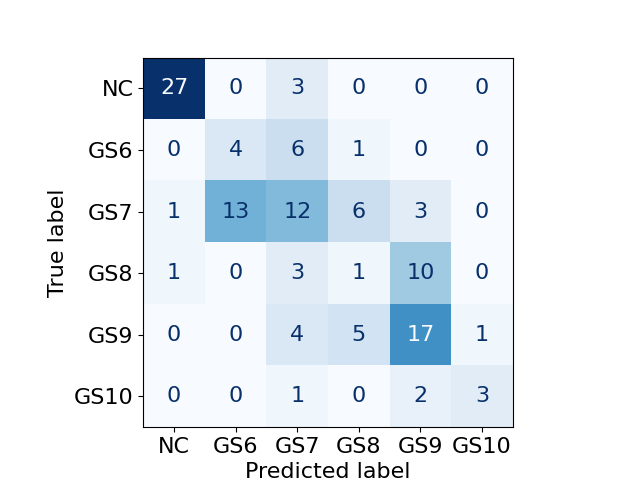}
    }
     \caption{Confusion matrices for the 4-fold cross-validation of SICAPv2 with the classes ``non-cancerous'' (NC) and Gleason Score 6 to 10 (GS6-GS10).}
     \label{fig:sicap_conf_mat}
\end{figure*}

\noindent
The SICAPv2 experiment is used to test our model on a very small dataset, where there is a high risk of overfitting. Recall that the training set for each cross-validation fold has less than 100 WSIs, and correspondingly there are less than 100 labels for the MIL models. 

As a first part of the SICAPv2 experiment, we perform an ablation study to show the effect of different hyperparameters that are important in the proposed attention module. 
We identify three hyperparameters that are especially interesting for this analysis: the dimension of the feature vectors $h$, the number of inducing points of the SGP, and the activation function inside the attention module (the one before the SGP).
We separately vary these hyperparameters while all the other values are set to default (as described in Section \ref{sec:prost_exp}, \textit{Implementation Details} paragraph).

As shown in Table \ref{tab:abl_study}, the AGP model is robust against variations of the feature vector dimensionality, but the model with $64$ feature dimensions slightly outperforms the others. 
For the number of inducing points we see a robust performance for less inducing points (16-32-64), but a high number (128) led to instabilities in model training. Indeed, we had to reduce the learning rate to 0.0001 (from 0.001) to obtain convergence, but we still observe a remarkable performance drop.
We believe that, as fully connected layers have reduced the complexity and dimensionality of the features, a relatively small amount of inducing points allows precise predictions.
Finally, the same convergence problems appear if a ReLu activation function is used before the SGP (we again used a lower learning rate of 0.0001 in this case to get convergence). The tanh function, which is more similar to the sigmoid function, shows a robust performance. 
Interestingly, these results suggest that limiting the input range to the SGP is clearly beneficial, as the sigmoid and tanh functions output values in the range $(0, 1)$ and $(-1, 1)$, respectively. The relu function has outputs in the range $[0, \infty)$, which makes it harder to find adequate inducing point locations for the SGP.

After the ablation study, the best performing model is compared to other state-of-the-art methods. 
Notice that the best performing configuration is precisely the one that was described in Section \ref{sec:prost_exp}, \emph{Implementation Details} paragraph.
The results in Table \ref{tab:comparison_sicap} show that the AGP model outperforms all other MIL approaches, including existing attention based methods A-Det and A-Det-Gated, which can be considered state of the art for MIL. Also, the Cohen's quadratic kappa value of $0.847$ is a remarkable one for such a small dataset. Furthermore, we see that AGP outperforms the existing supervised methods Silva-Rodríguez et al. \cite{silva-rodriguez_going_2020} and Arvaniti et al. \cite{arvaniti_automated_2018}, which use all patch-level annotations to train the feature extractor (reported by \cite{silva-rodriguez_going_2020} for this dataset). This can be partly explained by the stronger focus on patch-level predictions instead of bag-level predictions of these approaches (for bag-level predictions, they implement a simple aggregation method). Interestingly, notice that the AGP model provides an accurate WSI diagnosis without local annotations. 

We also report the confusion matrices for all the compared models, see Figure \ref{fig:sicap_conf_mat}. 
We see that AGP is strong in distinguishing cancerous from non-cancerous WSIs: there is only one false positive and one false negative in AGP predictions. The other two approaches that show a comparable performance in the binary (cancerous vs non-cancerous) classification task, Mean-Agg and A-Det-Gated, show a major systematic error: they misclassified all the WSIs with Gleason Score 10.

The training process of the proposed AGP model (with previously extracted features) takes less than 2 minutes for the SICAPv2 dataset, and is negligible in comparison to the training of the feature extractor ($\sim$ $7$ hours in this case \cite{schmidt_coupling}).
The test time is on average $2.1$ seconds for a complete WSI. This computing time mainly corresponds to the feature extraction ($\sim$ $7$ seconds \cite{schmidt_coupling}), while the attention mechanism and final classification together can be executed in less than $0.1$ seconds per WSI. The runtime bottleneck is therefore the feature extractor, and not the proposed probabilistic attention module. This efficiency is an important benefit for the clinical practice, as well as other areas where speed in prediction is paramount.

\subsection{PANDA Results}\label{sec:panda}

\begin{table}
\begin{center}

\caption{Results for PANDA dataset. We report the mean and standard error of Cohen's quadratic kappa ($\kappa$) for 4 independent runs. The last method does not report the standard error.}
\label{tab:comparison_panda}
\begin{tabular}{rccc}
\toprule
 Method      & Learning          & $\kappa$ mean & $\kappa$ S.E.\\
\midrule
Mean-Agg  & MIL &   0.803 & 0.003  \\
A-Det  & MIL &    0.811 &  0.004\\
A-Det-Gated  & MIL &   \textbf{0.816} &  0.004 \\
AGP  & MIL &    \textbf{0.817} & 0.003     \\ 
Silva-Rodríguez et al. \cite{silva-rodriguez_self-learning_2021}    & MIL        &      0.793     & N.A.\\
\bottomrule
\end{tabular}
\end{center}
\end{table}

\begin{figure*}
    \centering
     \subfloat[Silva-R. et al.]{
         \centering
         \includegraphics[trim={1cm 0.0cm 2.5cm 2cm}, width=0.22\linewidth]{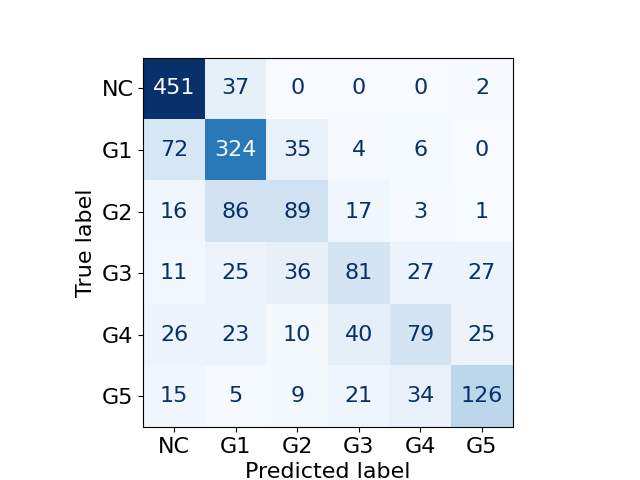}
    }     
     \subfloat[Mean-Agg]{
         \centering
         \includegraphics[trim={1cm 0.0cm 2.5cm 2cm}, width=0.22\linewidth]{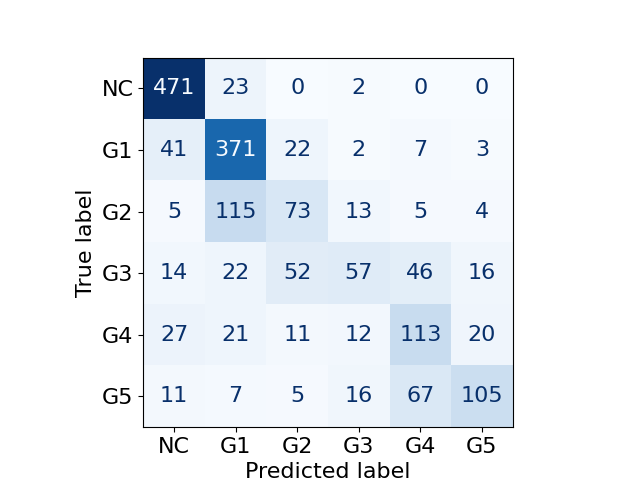}
    }      
    
    \bigskip
     \subfloat[A-Det]{
         \centering
         \includegraphics[trim={1cm 0.0cm 2.5cm 2cm}, width=0.22\linewidth]{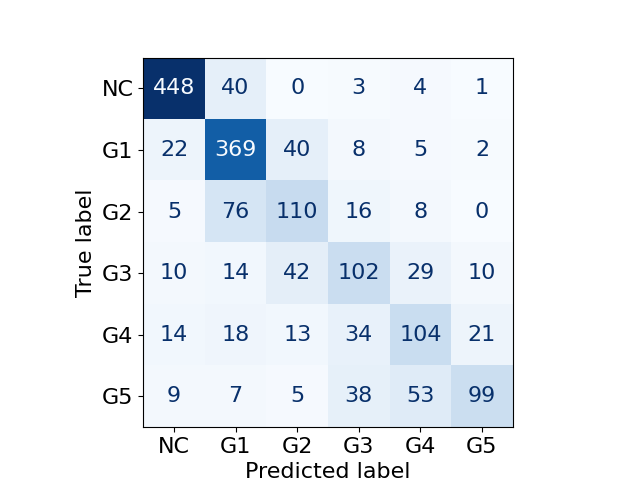}
    }     
     \subfloat[A-Det-Gated]{
         \centering
         \includegraphics[trim={1cm 0.0cm 2.5cm 2cm}, width=0.22\linewidth]{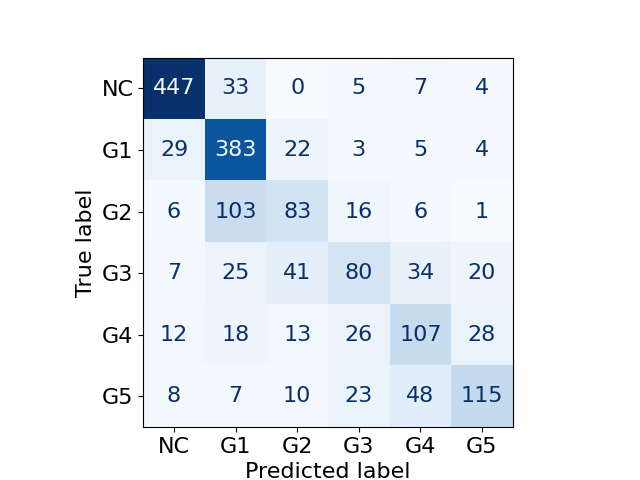}
    }  
     \subfloat[AGP]{
         \centering
         \includegraphics[trim={1cm 0.0cm 2.5cm 2cm}, width=0.22\linewidth]{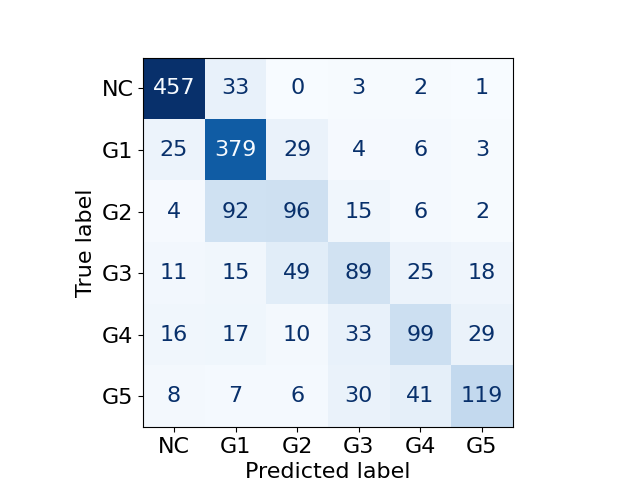}
    }  
     \caption{Confusion matrices for the PANDA test set. The six classes are ``non-cancerous'' (NC) and ISUP grades 1 to 5 (G1-G5). }
     \label{fig:panda_conf_mat}
\end{figure*}

\noindent
In this experiment, we show that AGP also outperforms other approaches in a large real-world problem. Moreover, by visually inspecting the predictions, we check that the AGP attention mechanism allows for identifying cancerous regions. Finally, we analyze the relevance of the probabilistic predictions provided by AGP, which can be used to detect wrong predictions.
Also, as explained before, the different grading scale used here (ISUP scale) shows the robustness of AGP.

In Table \ref{tab:comparison_panda} we see that the AGP performance is superior to the other MIL approaches. We have included Silva-Rodríguez et al. \cite{silva-rodriguez_going_2020}, a recent MIL method that was evaluated on the same dataset. Although the difference in performance is small in some cases (e.g. against A-Det-Gated), we will see that the probabilistic nature of AGP provides additional benefits (such as the degree of uncertainty on the predictions).

The confusion matrices, see Figure \ref{fig:panda_conf_mat}, show again that the AGP model is very strong at differentiating cancerous from non-cancerous WSIs. Although the models of Silva-Rodriguez et al. \cite{silva-rodriguez_going_2020} and Mean-Agg have less false positives (see the first row of the matrices), these models suffer from more false negatives (see first column of the matrices).
This can also be confirmed by the binary F1 score (cancerous vs. non-cancerous): AGP outperforms all other approaches with a binary F1 score of $0.960$ (Silva-R: $0.927$, Mean-Agg: $0.950$, A-Det: $0.958$, A-Det-Gated: $0.956$).

\begin{figure*}
     \centering
\includegraphics[width=\textwidth]{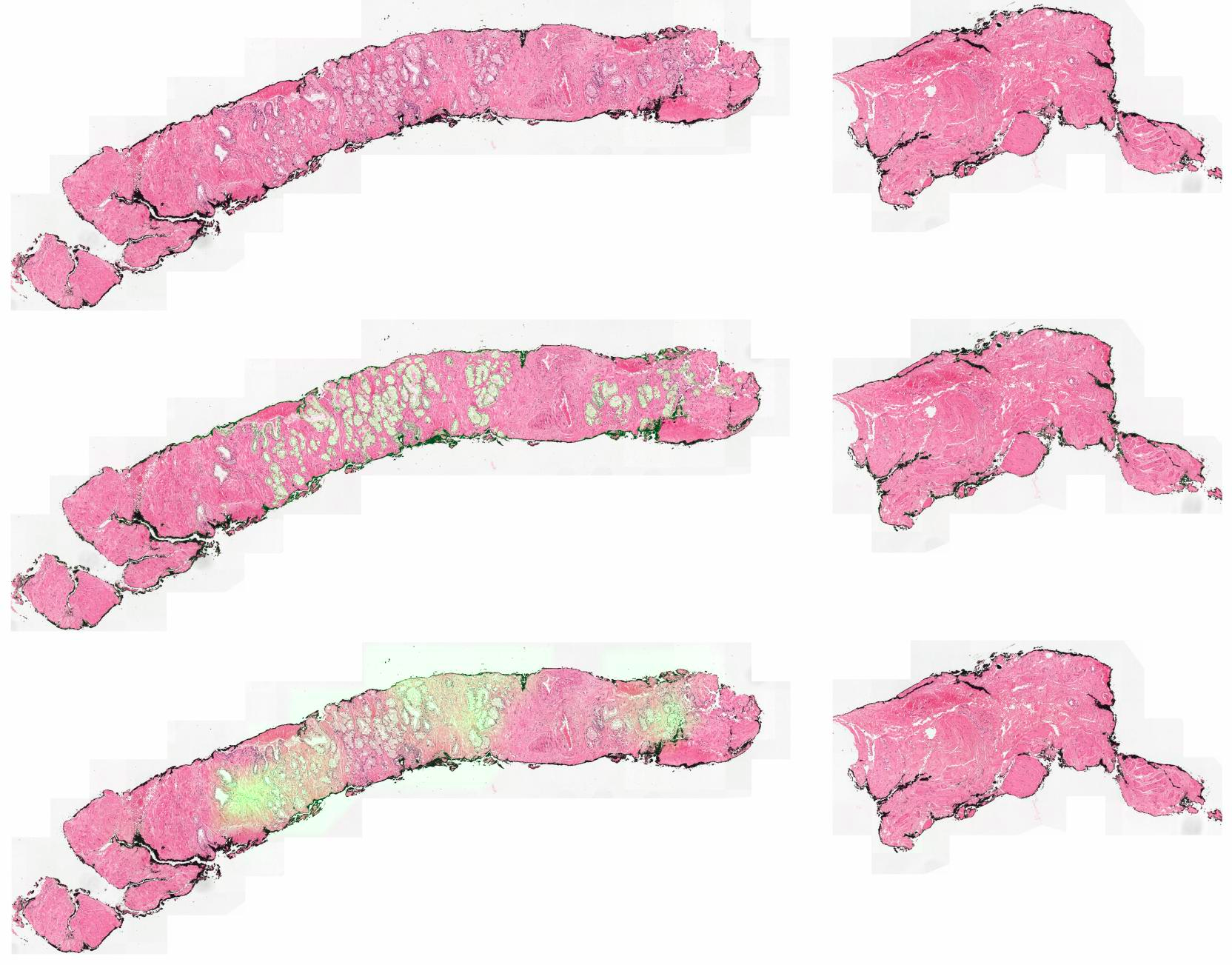}
\vspace{-7mm}
\caption{Attention weights for a test WSI of the PANDA dataset. The top image shows the original WSI, the middle image shows the cancerous areas marked by an expert pathologist (in green) and the bottom image shows the areas of high attention as predicted by the model (in green). The predicted attention weights were normalized and interpolated from the patch coordinates by linear interpolation. 
As can be seen in the image, the model successfully assigns high attention weights to the discriminative areas of the WSI. This improves explainability of the models prediction and helps the pathologist to find suspicious areas.}
\label{fig:wsi_att}
\end{figure*}

Next, we illustrate how the AGP attention mechanism provides an explainable prediction at instance level. The top row in Figure \ref{fig:wsi_att} shows one test WSI example (it has two pieces of tissue, a big one on the left and a small one on the right). In the second row, the cancerous areas are colored in green (this example has been manually segmented by an expert pathologist for this evaluation; recall that AGP only uses bag-level labels).
The third row shows the areas with high attention weights predicted by AGP highlighted in green. For this figure, the attention weights were predicted by the model for each patch of the image (remember that the patches are of 512x512 resolution with 50\% overlap). To obtain the heatmap for the complete WSI, linear interpolation was performed between the grid of patches.
We find that the parts with high attention correspond to areas of the WSI that are most affected by cancer. Other parts that are non-cancerous, such as the whole piece on the right, are not assigned high attention weights. This means that AGP works as expected and the final prediction is based on discriminative areas. The attention helps the pathologist verify the prediction and further inspect the affected tissue. Moreover, it might even point out cancerous regions that the pathologist may have missed.

\begin{figure*}
     \centering
\includegraphics[width=0.6\textwidth]{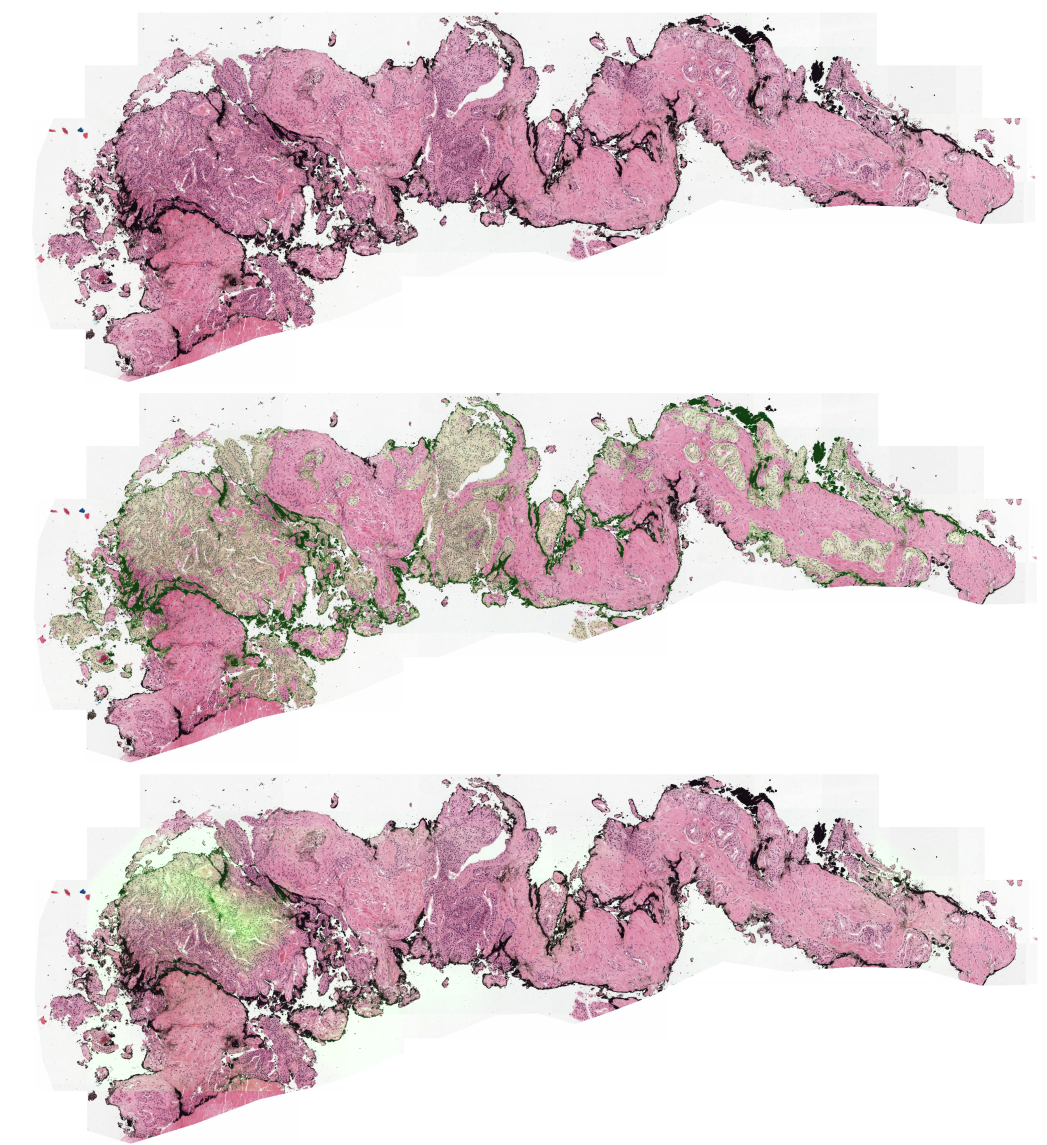}
\vspace{-3mm}
\caption{
Example of an inaccurate assignment of attention weights. 
The top image shows the original WSI, the middle image shows the cancerous areas marked by an expert pathologist (in green), and the bottom image shows the areas of high attention as predicted by the model (in green). 
Although in most cases the areas of high attention correspond to the tumorous areas, recall Figure \ref{fig:wsi_att}, we observed some inaccurate cases as the presented in this image. Here, the correct class (ISUP grade 4) is predicted, but the attention mechanism does not capture all the cancerous tissue parts.}
\label{fig:wsi_fail}
\end{figure*}

Although the attention mechanism works well for most WSIs (as supported by the superior predictive performance), we observed that for some WSIs the attention does not capture all the important areas. In Figure \ref{fig:wsi_fail} we show the same plots as in Figure \ref{fig:wsi_att} for a failure case of the attention mechanism (top: WSI, middle: annotation, bottom: attention estimation). In this case, not all the cancerous areas are assigned a high attention weight. However, the attention is still useful here as a source of explainability. It highlights all the areas which the classification is based on. Notice also that the highlighted areas are indeed tumorous, and the correct class (ISUP grade 4) is predicted.

\begin{figure}
\centering
\includegraphics[width=0.4\columnwidth]{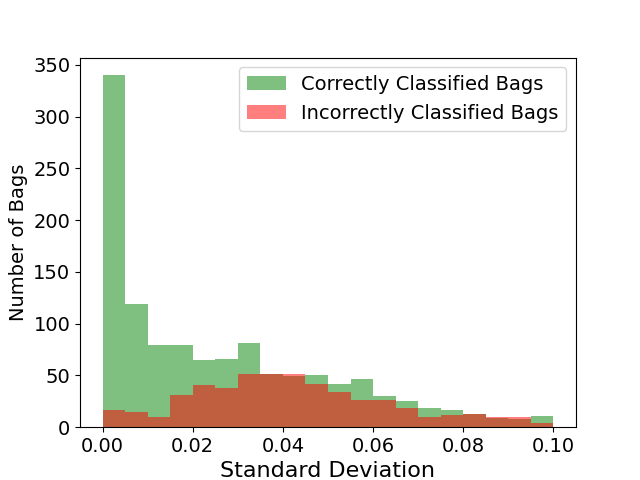}
\caption{Distribution of the predicted standard deviations. Bags (images) with a low standard deviation are likely to be classified correctly, while a high standard deviation indicates a high risk of a wrong prediction. The standard deviations are divided into bins of width 0.005, and the y-axis shows the number of bags with the corresponding std.}
\label{fig:std_distribution}
\end{figure}

Finally, we focus on the probabilistic bag predictions that provide not only a class score, given by the mean, but also the predictive uncertainty, given by the standard deviation.
Figure \ref{fig:std_distribution} shows a histogram over the predictive uncertainty (i.e. the standard deviation) for all the AGP bag predictions. The green (resp. red) bars indicate the number of correctly (resp. incorrectly) predicted bags whose standard deviation falls in a certain range. It is clearly visible that correct predictions tend to have a lower standard deviation than the incorrectly classified bags. In other words: a high standard deviation correlates with a high risk of a wrong classification. This suggests that the standard deviation provides a useful measure of the predictive reliability. In fact, if we only take the reliable predictions with a std. below $0.02$, the Cohen's kappa value for the PANDA test set rises to $0.864$ for the AGP model (from $0.817$ in Table \ref{tab:comparison_panda}). 
Therefore, in practice, the uncertainty estimation can help the pathologists decide when the models' prediction should be disregarded or double-checked.

\subsection{External Validation with PANDA and SICAPv2}\label{sec:extrapolation}

\noindent
The third experiment tests the generalization capability for the models evaluated in Section \ref{sec:panda}. 
Similar to \cite{silva-rodriguez_self-learning_2021}, we take the models trained on the PANDA dataset, and use all images of SICAPv2 as an external test set (since the training is done under ISUP grading, SICAPv2 uses ISUP grading here too; therefore, results are not comparable to those obtained in Section \ref{sec:sicap}).
Table \ref{tab:comparison_panda_train_sicap_test} shows the results.
Again, AGP outperforms the rest of approaches, and achieves a remarkable Cohen's quadratic kappa value of $0.92$. 

Similar to Section \ref{sec:panda}, it is worth analyzing the standard deviation of the predictions. In addition to distinguishing between correctly and incorrectly classified images, there is now an additional dimension to consider: whether the test images follow the same distribution as the training ones or not (i.e. whether we are using PANDA or SICAPv2 as test set, respectively). 
 
Table \ref{tab:std_external} shows the average predicted standard deviation of correct and wrong predictions for each of the test sets. The main findings are twofold: (1) the output standard deviation is higher for wrong predictions, and (2) the output standard deviation is higher for data originating from a different data distribution. This shows that the output uncertainty is not only helpful to identify model failures for in-distribution data, but also accounts for uncertainty added by a data shift. \cite{ovadia2019can}
 This can help to determine images that might be out of scope for the model and should be handled with caution.

\begin{table}
\begin{center}
\caption{Models trained on the PANDA dataset, tested on SICAPv2 with ISUP grading scale. The mean and standard error of Cohen's quadratic kappa ($\kappa$) are reported for four independent test runs. The last method does not report the standard error.}
\label{tab:comparison_panda_train_sicap_test}
\begin{tabular}{rccc}
\toprule
 Method      & Learning          & $\kappa$ mean & $\kappa$ S.E.\\
\midrule
Mean-Agg  & MIL &     0.911 & 0.007   \\
A-Det  & MIL &    0.903 & 0.004  \\
A-Det-Gated  & MIL &   0.910 & 0.007  \\
AGP  & MIL &    \textbf{0.920}  & 0.001   \\
Silva-Rodríguez et al. \cite{silva-rodriguez_self-learning_2021}      & MIL        &   0.885    & N.A.\\
\bottomrule
\end{tabular}
\end{center}
\end{table}

\begin{table}
\centering
\caption{The average standard deviation of correct and wrong bag predictions for AGP trained on the PANDA dataset. The first (resp. second) row shows the result when using PANDA (resp. SICAPv2) as test set. The values, which reflect the uncertainty in the prediction, get higher for wrong predictions and when testing on a different test set.\label{tab:std_external}}
\begin{tabular}{rcc}
\toprule
         & \multicolumn{2}{c}{Average uncertainty}\\
         \cmidrule[0.5pt](){2-3}
Test set & Correct predictions & Wrong predictions \\
\midrule
PANDA & 0.029 & 0.046 \\
SICAPv2 & 0.045 & 0.051 \\
\bottomrule
\end{tabular}
\end{table}

\section{Conclusions} \label{sec:conclusions}
We have proposed AGP, a novel probabilistic attention mechanism based on GPs for deep multiple instance learning (MIL). 
We have evaluated AGP in a wide range of experiments, including real-world cancer detection tasks. 
The novel attention module is capable of accurately assigning attention weights to the instances, and outperforms state-of-the-art deterministic attention modules. Furthermore, it provides important advantages due to its probabilistic nature. For instance, it addresses the problem of reliability in safety critical environments such as medicine: the probabilistic output of our model can be used to estimate the uncertainty on each prediction.
For future research, we plan to explore the use of deep GPs (instead of GPs) to further improve the performance.
Also, the promising results of AGP encourage the application of GP-based attention to other recent methods such as transformer networks, self-attention or channel attention.
Moreover, alternative probabilistic attention mechanisms based on other Bayesian approaches (instead of GPs) can be explored.

\bibliography{ms.bib}
\bibliographystyle{IEEEtran}
\textbf{Arne Schmidt} studied Mathematics and received his Bachelor degree at the Freie Universität Berlin in 2015 and his Master degree at the Technische Universität Berlin in 2018. 
He worked as a software engineer specialized in deep learning at Astrofein GmbH, Fraunhofer Heinrich Hertz Institut and TomTom. 
In 2020, he started his Ph.D. under the supervision of Prof. Rafael Molina at the University of Granada. It is part of the EU-funded H2020-MSCA-ITN CLARIFY, an interdisciplinary, international project on digital pathology for cancer classification.
His research interests are probabilistic deep learning, multiple instance learning, active learning and crowdsourcing with applications to medical images.

\textbf{Pablo Morales-Álvarez} received the B.Sc. degree in Mathematics from the University of Granada (UGR), Spain, in 2014. He obtained the First End of Studies Award by the Spanish Ministry of Science to the most outstanding undergraduate in mathematics in Spain. Then, he received the M.Sc. degrees in Mathematical Physics (2015) and
Data Science (2016), both from UGR. Funded
by the highly competitive Ph.D. fellowship from
La Caixa Banking Foundation, he got his Ph.D. degree in 2020 at UGR under the supervision of Prof. Rafael Molina (UGR) and Prof. Aggelos K. Katsaggelos
(Northwestern University, USA).
During his PhD he visited several research groups, including the Machine Learning Group at the University of Cambridge (UK), where he worked for five months with Prof. José Miguel Hernández-Lobato. 
From October 2020 to May 2021, he did postdoctoral research at Microsoft Research Cambridge (UK) with Cheng Zhang.
He has published in top machine learning conferences and journals, such as ICLR 2021, NeurIPS 2022 and IEEE Trans. PAMI. He has also served as a reviewer for top-tier conferences (ICML 2020, NeurIPS 2020, ICLR 2021, NeurIPS 2021), and journals (Science).
He has obtained the SCIE-FBBVA Award 2022 to the Most Outstanding Young Spanish Researchers
in Computer Science.
His main research interests are probabilistic machine learning methods, specially (Deep) Gaussian Processes and Bayesian Neural Networks.

\textbf{Rafael Molina}
received
the M.Sc. degree in mathematics (statistics) and the Ph.D. degree in optimal design in linear models from the University of Granada, Granada, Spain,
in 1979 and 1983, respectively. He was the Dean of the Computer Engineering
School, University of Granada, from 1992 to 2002,
where he became a Professor of computer science
and artificial intelligence in 2000. He was the Head
of the Computer Science and Artificial Intelligence
Department, University of Granada, from 2005 to
2007. 
He has coauthored an article that received the runner-up prize at
reception for early stage researchers at the House of Commons in 2007.
He has coauthored an awarded Best Student Paper at the IEEE International
Conference on Image Processing in 2007, the ISPA Best Paper in 2009, and
the EUSIPCO 2013 Best Student Paper. His research interest focuses mainly
on using Bayesian modeling and inference in image restoration (applications
to astronomy and medicine), super-resolution of images and video, blind
deconvolution, computational photography, source recovery in medicine,
compressive sensing, low-rank matrix decomposition, active learning, fusion,
supervised learning, and crowdsourcing.
Dr. Molina has served as an Associate Editor for Applied Signal Processing
from 2005 to 2007 and the IEEE TRANSACTIONS ON IMAGE PROCESSING
from 2010 to 2014. He has been serving as an Area Editor for Digital Signal
Processing since 2011.

\end{document}